\documentclass[letterpaper]{article} 
\usepackage{aaai23}  
\usepackage{times}  
\usepackage{helvet}  
\usepackage{courier}  
\usepackage[hyphens]{url}  
\usepackage{graphicx} 
\usepackage{natbib}  
\usepackage{caption} 
\usepackage{todonotes}
\usepackage{xspace} 
\usepackage{multirow}
\usepackage{amsmath}
\usepackage{amsfonts}
\usepackage{xcolor}
\usepackage{algorithm}
\usepackage{algorithmic}

\usepackage{newfloat}
\usepackage{listings}
\DeclareCaptionStyle{ruled}{labelfont=normalfont,labelsep=colon,strut=off} 
\floatstyle{ruled}
\newfloat{listing}{tb}{lst}{}
\floatname{listing}{Listing}
\title{SafeLight: A Reinforcement Learning Method toward Collision-free Traffic Signal Control}
\author{
    Wenlu Du,
    Junyi Ye,
    Jingyi Gu,
    Jing Li,
    Hua Wei,
    Guiling Wang\footnote{Corresponding Author.}
}
\affiliations{
  New Jersey Institute of Technology, Newark, USA\\
    \{wd48, jy394, jg95, jingli, hua.wei, guiling.wang\}@njit.edu
}

\newcommand{\methodFont}{\textsl}
\newcommand{\formula}{\methodFont{Fixed-time}\xspace}
\newcommand{\singleobjective}{\methodFont{Backbone RL(3DQN)}\xspace}
\newcommand{\syntheticreward}{\methodFont{Syn-R}\xspace}
\newcommand{\syntheticvalue}{\methodFont{Syn-Q}\xspace}
\newcommand{\Safelight}{\methodFont{SafeLight}\xspace}
\newcommand{\SafelightAct}{\methodFont{SafeLight-Act}\xspace}
\newcommand{\SafelightR}{\methodFont{SafeLight-R}\xspace}
\newcommand{\SafelightLoss}{\methodFont{SafeLight-Loss}\xspace}
\newcommand{\SafelightSR}{\methodFont{SafeLight-S\&R}\xspace}
\newcommand{\SafelightSLoss}{\methodFont{SafeLight-S\&Loss}\xspace}

\usepackage{bibentry}

\begin{document}

\maketitle

\begin{abstract}
Traffic signal control is safety-critical for our daily life. Roughly one-quarter of road accidents in the U.S. happen at intersections due to problematic signal timing, urging the development of safety-oriented intersection control. However, existing studies on adaptive traffic signal control using reinforcement learning technologies have focused mainly on minimizing traffic delay but neglecting the potential exposure to unsafe conditions. We, for the first time, incorporate road safety standards as enforcement to ensure the safety of existing reinforcement learning methods, aiming toward operating intersections with zero collisions. We have proposed a safety-enhanced residual reinforcement learning method (\Safelight) and employed multiple optimization techniques, such as multi-objective loss function and reward shaping for better knowledge integration. Extensive experiments are conducted using both synthetic and real-world benchmark datasets. Results show that our method can significantly reduce collisions while increasing traffic mobility.
\end{abstract}

\section{Introduction}
Traffic congestion has become increasingly costly. For example, on average, American drivers lost 26 hours in traffic jams even in 2020 under the pandemic situation, which already includes a drop from 73 hours in 2019~\cite{inrix}. The American Transportation Research Institute estimates that congestion costs the U.S. freight sector \$74.1 billion annually, of which \$66.1 billion is from urban areas~\cite{glover2020}. Intersections, especially in the urban area, are where the aforementioned congestion problems happen every single day. Signalized intersections are one of the most common bottleneck road types in urban environments. Thus, Traffic Signal Control (TSC) plays a vital role in urban traffic management~\cite{mcelroy2007congestion}.

To improve the mobility of vehicles in a city, Reinforcement Learning (RL), a widely adopted method for control problems, has been applied to TSC research and has proven to be effective~\cite{genders2016using,zheng2019learning,chen2020toward,wei2021recent}. The biggest advantage of RL is that it directly learns how to take the next action by observing the feedback (rewards) from the environment after previous actions and continuously explores the possible cases. By setting mobility-related rewards like vehicle delay, RL-based TSC methods can learn to adjust the traffic signals to accommodate the real traffic demand and hence achieve better than human performance to make traffic move faster.

However, TSC must consider both traffic mobility and traffic safety. A small mistake in traffic signals can cause significant loss of life and property. According to the Federal Highway Administration, more than 50 percent of the combined fatal and injury crashes occur at or near intersections~\cite{coben2006national}. Moreover, crashes at signalized intersections occupy about one–third of all intersection fatalities, including a large proportion that involves red-light running~\cite{himes2017safety} and improper left turns~\cite{le2018safety}, which could be reduced through well-designed signal timing plans. This motivates the study of safety rules in the traditional transportation field.

Conventional transportation research has extensively studied the safety effects associated to signal timing. In particular, previous works~\cite{anjana2015safety,wong2007contributory}  analyzed the contributory factors for traffic crashes and found red-light violations, yellow-light time changes, and left-turn phases are critical for traffic safety. Other works~\cite{essa2019full,Persaud2002} predicted the safety risk through regression models and Bayesian methods and provided authoritative equations and rules~\cite{urbanik2015signal} as domain knowledge to help traffic engineers determine the signal timing in practice.

Some RL-based TSC methods attempted to address the safety concerns. They usually incorporate safety into the reward and combine it with mobility-related rewards~\cite{khamis2014adaptive,Gong2020,wei2018intellilight}. Unfortunately, this often results in highly sensitive performance w.r.t. the setting and inevitably leads to a long learning process. 
In other words, finding the appropriate rewards of traffic mobility and safety for RL models requires not only the ad-hoc tuning of various reward factors but also a long time of exploration to learn the safe actions. The tuning process can be avoided by Inverse Reinforcement Learning~\cite{ng2000algorithms} where the reward signal is learned by the RL agent, but still it would require a large volume of data to learn the reward for safety, large enough to cover various unsafe cases which are difficult to obtain. The long training time is painful especially since accidents caused by actions of traffic signals are rare during training, RL methods have to explore a large number of cases to learn the safety rules. Moreover, even the well-learned model might not obey the safety rules in practice.

Therefore, can we better incorporate safety rules and improve RL to address both traffic safety and mobility requirements? Such a safety-enhanced RL model is a critical step closer to the real-world deployment of RL-based TSC methods. In this work, we step toward this goal and show that frequent traffic crashes can be avoided through carefully designed RL-based TSC methods. By investigating the available domain safety rules, we formulate a safety model accordingly to decide if a TSC action is unsafe. Next, we incorporate the safety model as add-on safety modules into the RL model with systematic analysis. More precisely, we integrate it into the different parts of the RL design, including the state, action, reward and loss function. A safety-enhanced RL approach, \Safelight, which utilizes residual learning~\cite{Johannink2018}, is then proposed. We analyze the pros and cons of each design choice for the safety modules and conduct experiments on both synthetic and real-world datasets from existing benchmarks. Results show that our methods can be easily integrated into existing works and achieve superior performance under both safety and mobility measures. 

In summary, the contributions of this work are as follows:

$\bullet$ 
We reveal that current RL-based methods can impose safety concerns. Compared with conventional non-RL methods, the occurrences of collisions are higher in both synthetic and real-world benchmark datasets. 

$\bullet$ 
We provide a systematic analysis of possible integration of domain safety rules and propose \Safelight. Through testing on two RL models with different designs, we verify that our residual-based method can be incorporated into existing RL models, or most RL methods in general, to enhance their safety to near zero-collision level. 

$\bullet$ 
We evaluate our proposed methods on multiple datasets. Results show that \Safelight achieves over 99 percent reduction in collisions than the backbone RL model and about 30 percent lower average waiting time than the \formula control. The excellent performance of our proposed method holds under different environment settings.

\section{Related Works}

\subsection{RL-based TSC Methods}
There have been many research works concentrated on improving traffic mobility using RL algorithms. This is largely because they can adapt to real-time traffic, which directly evolves in complex and unpredictable circumstances, such as real geometric road conditions and unexpected accidents and congestion \cite{khattak1991driver} \cite{wang2014multi}. Those that claim the state-of-the-art performance in TSC can be roughly categorized into two groups: 1) Deep Q-Network (DQN) based RL algorithms~\cite{mnih2015human} where the neural network is used to decide the action taken by an RL agent, and 2) Actor-Critic Network based RL algorithms~\cite{mnih2016asynchronous} where an extra neural network is used to estimate the state advantages. Among the first group, a Double DQN (DDQN) with a dual-agent algorithm~\cite{van2016deep} is proposed to obtain a stable traffic signal control policy. The work of \cite{zhang2021independent} extends the DDQN algorithm using the forgetful experience mechanism and the lenient weight training mechanism to speed up training. A Dueling DDQN (3DQN) model~\cite{Liang2019} is proposed with prioritized experience replay to further improve the sampling efficiency. In the second group, Proximal Policy Optimization (PPO)~\cite{schulman2017proximal}, which follows the advantage actor-critic paradigm, is employed in the work of \cite{ault2019learning} to achieve a smooth and monotonic learning curve. However, these RL methods only improve mobility without considering safety concerns.

There are a few RL-based works that consider multiple objectives, including safety. To be deployed in the real world, traffic signals must provide both mobility and safety~\cite{koonce2008traffic}. In \cite{khamis2014adaptive}, a multi-agent RL method with a synthetic reward is proposed to optimize multiple objectives, i.e., reducing trip waiting time, total trip time, junction waiting time and collision risks. The main issues of this method are: (1) the weights of the rewards for different objectives need to be determined by domain experts; and (2) the convergence may not be guaranteed. Instead of using synthetic rewards, some works~\cite{jin2015adaptive,Gong2020} consider modeling the synthetic value function or synthetic Q-values (i.e., the synthetic expected long-term reward) for improving both mobility and safety. This method is also known as multi-objective RL (MORL), which learns different value functions independently and then makes the decision based on the synthetic Q-values. Compared to synthetic reward-based models, MORL methods guarantee to converge. The main concern is that the traffic safety risk estimation in their framework requires detailed labeled traffic crash data in the study area, which is usually not available for most roads.

\subsection{Safe Reinforcement Learning}
In high-stakes domains like transportation, safety is particularly important, thus researchers are paying attention to both long-term reward and safety violation avoidance. Based on the survey paper~\cite{garcia2015comprehensive}, two main strategies for Safe Reinforcement Learning are considered: one is the modification of the optimality criterion and the other is the modification of the exploration process to avoid unsafe situations. The former, for example in the work~\cite{di2012policy}, uses constrained criterion to maximize the return while keeping other expected measures within some certain bounds; while the latter takes advantage of external knowledge, such as teacher advice~\cite{geramifard2012practical}. Others~\cite{thomas2021safe} \cite{alshiekh2018safe} propose Safe Reinforcement Learning by planning ahead a short time into the future to anticipate safety violations before they occur. Safe RL is a promising path toward applying RL to safety-critical problems yet has not been applied to TSC.

\section{Background}
This section will describe preliminary knowledge about reinforcement learning (RL). Moreover, background knowledge about current safety rules and standards in the transportation field will be introduced.

\subsection{Reinforcement Learning}\label{sec:3dqn}
An RL agent learns decisions through interactions with the environment that is modeled as a Markov Decision Process (MDP), defined as a tuple $(S,A,P,R,\gamma)$, where $S$ denotes a state space, $A$ denotes an action space, $P: S\times A\times S \rightarrow [0,1]$ denotes transition probabilities between states, $R$ denotes a reward function, and $\gamma \in [0,1]$ is a discount factor. Specifically, at each time period $t$, the agent observes a state $s_t \in S$ and takes an action $a_t\in A$, which is determined by the policy $\pi:S\rightarrow A$. Then, the next state $s_{t+1}$ is reached with a transition probability $T(s_{t+1}|s_t,a_t)$, and the agent receives a reward $r_t\in R$. The action-value function $Q^\pi$ is defined to evaluate how good it is for an agent to pick the action $a_t$ based on policy $\pi$ in state $s_t$. It is expressed as the expected cumulative reward: $Q^\pi (s,a)=\mathbb{E}[\sum_{i=t}^{\infty}\gamma^i r_{t+i}|s_t=s,a_t=a]$. The objective of an RL agent is to learn the optimal policy $\pi^*$ for maximizing $Q^\pi (s,a)$.

\subsection{TSC as an MDP}
When RL is applied in TSC, the signalized intersection environment is commonly modeled as the following MDP.

\subsubsection{State Space} 
The controller receives a representation of the current intersection state. In literature, different factors are considered in the state space definition, e.g., the number of stopped vehicles, the number of approaching vehicles (i.e., queue length), the average speed of approaching vehicles, and the maximum of the vehicle waiting time. Some papers also use the vehicle positions with the help of advanced sensing capabilities. The state space can be any subset of the above sensing data, among which queue length is the one commonly used in most existing works.

\subsubsection{Action Space} 
\begin{figure}[t!]
  \centering
  \includegraphics[width=\linewidth]{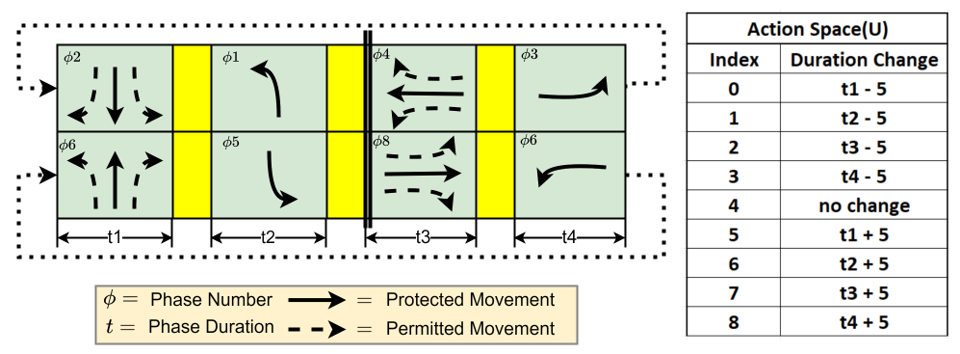}
  \caption{Illustration of the cyclic action space.}
  \label{fig:action}
\end{figure}

Based on the state, the controller chooses an action (i.e., signaling decision) to take. There are two major types of action space in TSC: cyclic and acyclic action spaces. In the \textbf{cyclic} setting, the control action is defined as a cycle --- a complete sequence of signal phases with computed duration. As shown in Figure~\ref{fig:action}, the cycle has four phases in the order of $(\phi2, \phi6),(\phi1, \phi5), (\phi4, \phi8), (\phi3, \phi6)$. The RL agent interacts with the environment cycle by cycle, adaptively adjusting a phase duration in the cycle by either increasing or decreasing its current duration. In the \textbf{acyclic} setting, the control action is defined as a phase $p$. Each time the agent chooses a phase $p$ from all possible phases as its action $a_i^t$ to take at the current state. In this setting, the execution time of each action, $\Delta t$, is fixed.

\subsubsection{Reward Function} 
After taking an action, the traffic controller receives a reward to indicate how good the signaling decision was. The reward function is one important signal to guide the RL agent toward the well-defined objective, e.g., improving mobility. Vehicle delay, waiting time, intersection throughput, and queue length can also be used in the reward function.

\subsection{Domain Strategy on Safety}
Safety strategies from the transportation domain are in the form of logical rules and traffic guidelines from authorities. Safety around an intersection is affected by many factors, including signal control, intersection geometry, traffic demand/volume, driving behavior, etc. Among all factors, signal control is the crucial one that traffic engineers take charge of. Transportation authorities like the Institute of Transportation Engineers (ITE) offer authoritative domain equations for calculating signal timing parameters with safety aspects in consideration. In addition, the Department of Transportation also provides safety-conscious \textit{design standards}~\cite{traffic2009manual, wolshon2016traffic, lockwood1997ite, koonce2008traffic}. In this subsection, we use the left-turn phasing in the domain design standards as an example to illustrate the potential safety concerns. For other standards like yellow change and red clearance interval, we refer to~\cite{urbanik2015signal}.

\subsubsection{Logical Rules for Left-turn Phasing} 
\begin{figure}[h]
  \centering
  \includegraphics[width=.35\textwidth]{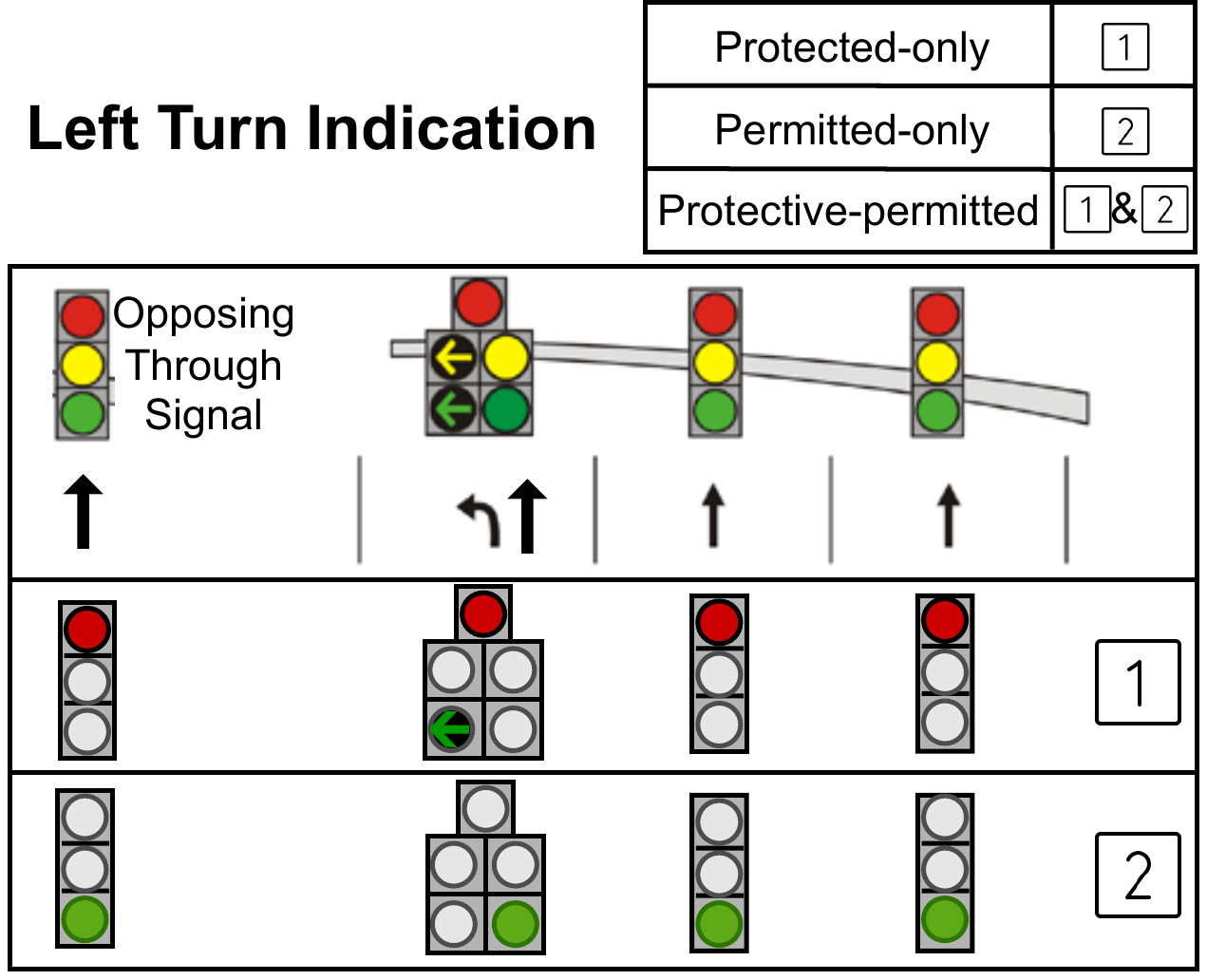}
  \caption{Illustration of protected, permitted and protective-permitted left-turn phasing mode with 3 approaching lanes. For protective-permitted mode, the protected and permitted left turn phases coexist in one cycle.}
  \label{fig:left-turn}
\end{figure}

As is shown in Figure~\ref{fig:left-turn}, left-turn phasing in practice often operates in the following modes: \emph{protected-only}, where left-turn drivers have the exclusive right of the way, \emph{permitted-only}, where left-turn drivers may yield to the opposing vehicles, and \emph{protective-permitted} which is the combination of the aforementioned two. Many studies have shown that \emph{permitted-only} and \emph{protective-permitted} operations can reduce delay while having a negative impact on safety. 
To ensure safety, it is possible to adjust the left-turn phasing mode by integrating domain rules or guidelines. Several left-turn phasing guidelines in the Federal Signal Timing Manual~\cite{koonce2008traffic} could be used to build our safety model, which determines if a left-turn phasing mode is unsafe under certain conditions. For example, based on the guidelines, if the $85\%$ percentile speed~\cite{koonce2008traffic} of opposing traffic is greater than $45$ mph, then the permitted left-turn phasing mode is unsafe. More details of the logical rules used in our design can be found in the Appendix. 

Note that there are different safety rules or guidelines on left-turn phasing. In this paper, we refer to the guidelines suggested by the Federal Signal Timing Manual~\cite{koonce2008traffic}, and other guidelines could also be similarly integrated into our proposed framework. 

\section{Proposed Approach}
\begin{figure*}[t!]
  \centering
  \includegraphics[width=.98\textwidth]{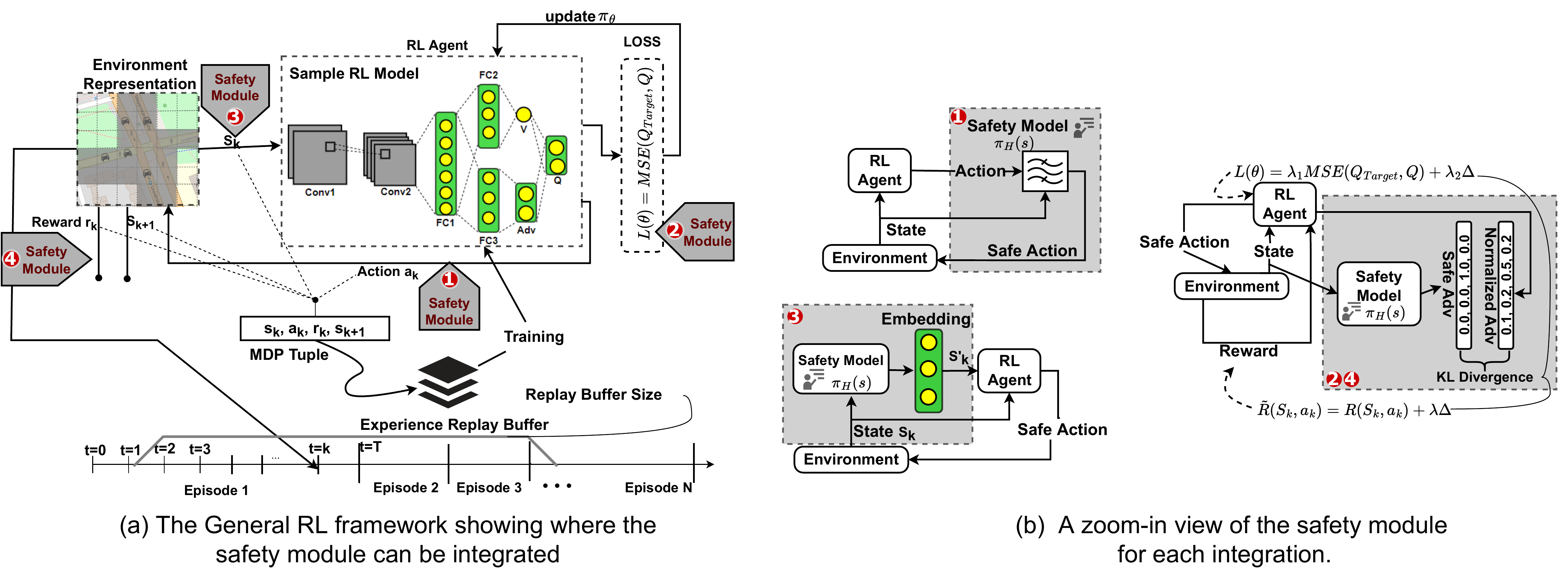}
  \caption{RL framework with four possible ways of integrating safety modules in action, state, reward function, and loss function. Note that the RL model could be in any form, while this figure only illustrates one example model adopted from~\cite{Liang2019}.}
  \label{fig:overall_structure}
\end{figure*}

This section introduces the RL model and the proposed safety-module integration framework \Safelight with a comprehensive discussion of the pro and cons of its variations. 

\subsection{RL Model}
We implement our proposed safety-module integration approach based on an existing RL model~\cite{Liang2019} that has the state-of-the-art performance. In this model, sensing devices, such as vehicle-tracking cameras, are assumed to track the position and speed of vehicles within a certain sensing radius. The raw traffic data is pre-processed to image-alike matrices by segmenting the intersection into same-size square-shape cells with a two-value vector $\langle position, speed\rangle$ of the vehicles if any. To showcase that \Safelight is independent of RL design settings, both cyclic actions from \cite{Liang2019} and acyclic actions from another benchmark RL model~\cite{ault2019learning} are tested in our experiments. The reward function is the cumulative vehicle waiting time between the actions measured from the point of time when the vehicles have entered the intersection until the start of the green light. 3DQN is chosen as the RL algorithm.

\subsection{Safety Module Integration: \Safelight}
The proposed \Safelight for integration, see Figure~\ref{fig:overall_structure}, is described here. The safety modules include a safety model, $\pi_{H}(s)$, which takes the current traffic condition as input and verifies whether an action is unsafe. To avoid the turn crash in our example case, the safety model consists of the logical rules abstracted from the left-turn phasing guidelines. For instance, if a left turn is permitted but not protected and the foe vehicles are approaching the junctions at a faster speed, then the left turn should be switched to protected mode. More details can be seen in the Appendix.

We describe below how the safety modules can be integrated into different parts of a general RL framework.

\subsubsection{\SafelightAct} 
A safety module can be integrated into RL by directly influencing the actions. Inspired by the residual RL method~\cite{Johannink2018}, where the final action comes from the action learned from an RL agent and the action from a human-designed controller, in our setting, the action from the RL agent is filtered by the safety model as shown in Figure ~\ref{fig:overall_structure}(b) with a mark of $1$. Instead of directly learning a safe action, the learned action is corrected whenever a potential safety issue is detected by the safety guidelines. Such supervision is partial (a minimum interference only if the action is unsafe). For example, when the action given by RL includes a permitted left turn and it is determined unsafe, the permitted left turn signal will turn red, prohibiting all left-turn movements.

\noindent\textsc{Analysis}
~\textbf{Pros}: First, it decouples the goal of improving mobility and safety. The RL model is only for improving mobility, while the safety module is only used as a proactive measure for any safety concern and corrects the RL agent only if necessary. This is especially effective when there are only a few conflicting movements, e.g., nighttime. Moreover, the safety module can be easily built upon the RL model with minimal modifications to the RL model. Last, the final executed action filtered by the safety module can guarantee safety since the safe action is not learned by the RL agent. \textbf{Cons}: It highly depends on the current volume of conflicting vehicles. If the RL agent is corrected by the safety module frequently, the training process will fluctuate immensely and be hard to converge.

\subsubsection{\SafelightLoss} 
The safety module can also be integrated into the loss function, which becomes multi-task learning. In multi-task learning, where the performance relies on multiple optimization objectives, the standard approach is to train a model that can minimize a loss function including multiple terms $\mathcal{L}(.,.,\lambda)$ parameterized by a vector $\lambda$, corresponding to different optimizing tasks~\cite{kokkinos2017ubernet,zamir2018taskonomy}. Simple average or weighted sum are two common methods to balance different terms: $\mathcal{L}(. , ., \lambda) = \sum_i\lambda_i\mathcal{L}^i(. , .)$. We leverage such knowledge to build upon the existing RL model layering up one more optimizing task, i.e., minimizing collisions, into the loss function as shown in the equation below.

\begin{gather}
    \nonumber \mathcal{J} =\lambda_1\sum\limits_s\mathcal{P}(s)[Q_{target}(s,a) - Q(s,a;\theta]^2 + \lambda_2 \Delta, \\
    \nonumber\Delta = \mathcal{D}_{KL}(\hat{A}(\pi_H(s))||A(s,a;\theta))
\end{gather}
where $Q_{target}= r + \gamma Q(s', arg\max_{a'}(Q(s',a';\theta)),\theta^-)$ is generated by double Q-learning algorithm  and  $A(s,a;\theta) = Q(s, a) - V(s)$ is the advantage function. The first term of the loss function is from the backbone RL algorithm (3DQN), by which the parameters can be updated by the Mean Square Error (MSE) of the target network and online network. $\Delta$ denotes the second term as illustrated in Figure~\ref{fig:overall_structure}(b) with a mark of $2$, advocating safer actions through minimizing the gap between the actual action and the related safe action. Specifically, the Kullback-Leibler (KL) divergence~\cite{hershey2007approximating} of the normalized advantages over the action space and the desired safe advantage distribution inferred by the safety model $\pi_H(s)$ is measured. The normalized advantages reflect the agent's current preference over the actions and can be used as a measure of how far it is to the desired safe action. Note that when the action from the RL agent is safe (i.e., no collisions occur), the value of the second term in the loss function will be zero.

\noindent\textsc{Analysis}
\textbf{Pros}: The merit comes from the idea of multi-task learning, with one model optimized towards two objectives.
\textbf{Cons}: Such integration needs to fine-tune the coefficient factor of the two optimization objectives in the loss function for each intersection-specific scenario. Domain scientists in the transportation field still have to develop an understanding of the backbone RL algorithm before being able to set the coefficients reasonably.

\subsubsection{\SafelightR} 
A safety module can be integrated into the reward function. Similar to the idea of reward shaping~\cite{ng1999policy}, the domain knowledge is transformed as additional rewards to guide the RL agent to incorporate human expertise. We consider a general form of reward shaping, i.e., the additive form formally defined as $r' = r + F$, by adding a safety-related numeric reward value. In our approach, we take the same strategy as how we integrate it into the loss function, shown in Figure~\ref{fig:overall_structure}(b) with a mark of $4$. Once an unsafe action is confirmed by the safety model, a desired safe advantage distribution is formulated. Then the difference between the true advantage distribution over the action space and the desired safe advantage distribution can be measured using the KL divergence and used as an extra numeric reward value denoting the penalty for deviating from desired safe standard, i.e.,
\begin{equation}
    \nonumber\tilde{R}(S_k, a_k)= R(S_k, a_k) + \lambda\mathcal{D}_{KL}(\hat{A}(\pi_H(s))||A(s,a;\theta))
\end{equation}

\noindent\textsc{Analysis.}
~\textbf{Pros}: Owning to the reward-driven property of RL, modifying the reward is an effective and straightforward way to help the RL agent find an optimal solution considering both objectives.
\textbf{Cons}: It may suffer from reward shaping, where the performance highly relies on the design of the reward function. Since the estimated policy from DNN does not have a one-to-one mapping to the multiobjective reward, there is no theoretical proof for convergence.

\subsubsection{\SafelightSR and \SafelightSLoss} 
A safety module can be integrated into the state. The safety model can be embedded into an m-dimensional latent space via a layer of Multi-Layer Perceptron, as shown in the middle of Figure~\ref{fig:overall_structure}(b) with a mark of $3$. We define it as $Embed(\pi_H) = \sigma(\pi_H W_e + b_e)$ where $W_e$ and $b_e$ are weight matrix and bias vector to learn, and $\sigma$ is ReLU function. This embedding is then incorporated into the Q network via concatenation to serve as an additional feature. Such feature-rich input can be combined with \SafelightLoss or \SafelightR to further boost the performance.

\noindent\textsc{Analysis.}~\textbf{Pros}: It is a feature-rich method. The safety model is processed to a feature and encoded into the state representation. If used properly, we may obtain a performance gain since all DNN methods require a well-developed input representation. \textbf{Cons}: It must be combined with \SafelightLoss or \SafelightR methods. Otherwise, without a safety-aware performance measure during training, the RL model cannot learn from this safety embedding.

\section{Experiments}
In this section, we conduct experiments to answer the following research questions\footnote{The source codes are publicly available from https://gitlab.com/wenlu057/traffic-safety.git.}: 

\noindent$\bullet$
\textbf{RQ1}: Compared with state-of-the-arts, would the \Safelight outperform under both mobility and safety measures (i.e., reducing collisions without noticeably compromising mobility)?

\noindent $\bullet$
\textbf{RQ2}: Can \Safelight maintain the same level of performance under different RL designs (i.e., cyclic and acyclic actions)?

\noindent $\bullet$
\textbf{RQ3}: Is \Safelight flexible to be integrated into different backbone RL models (i.e., 3DQN and IPPO)? 

\subsection{Evaluation Environment}
The experiments are conducted in the microscopic traffic simulator, SUMO, under time-variant traffic flow. 
\begin{figure}[ht]
  \centering
  \includegraphics[width=\linewidth]{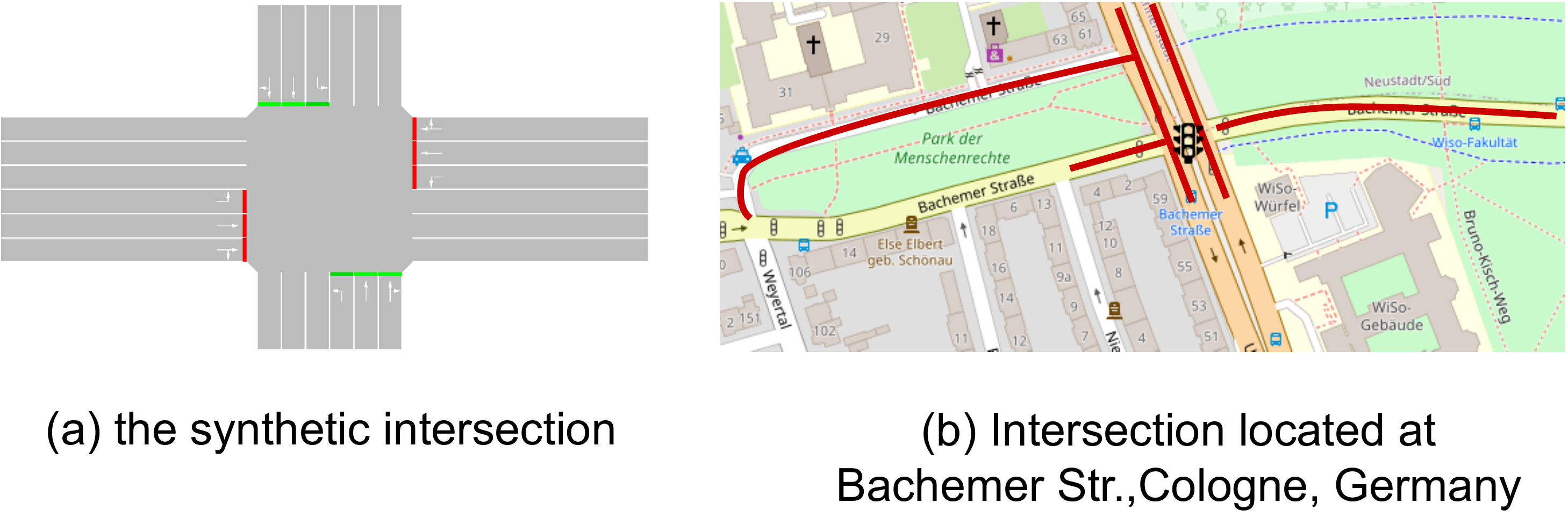}
  \caption{(a) the intersection geometry for synthetic benchmark dataset. (b) the real and complex intersection geometry for a real-world benchmark dataset.}
  \label{fig:dataset}
\end{figure}
\begin{figure*}[!t]
  \centering
  \includegraphics[width=.99\textwidth]{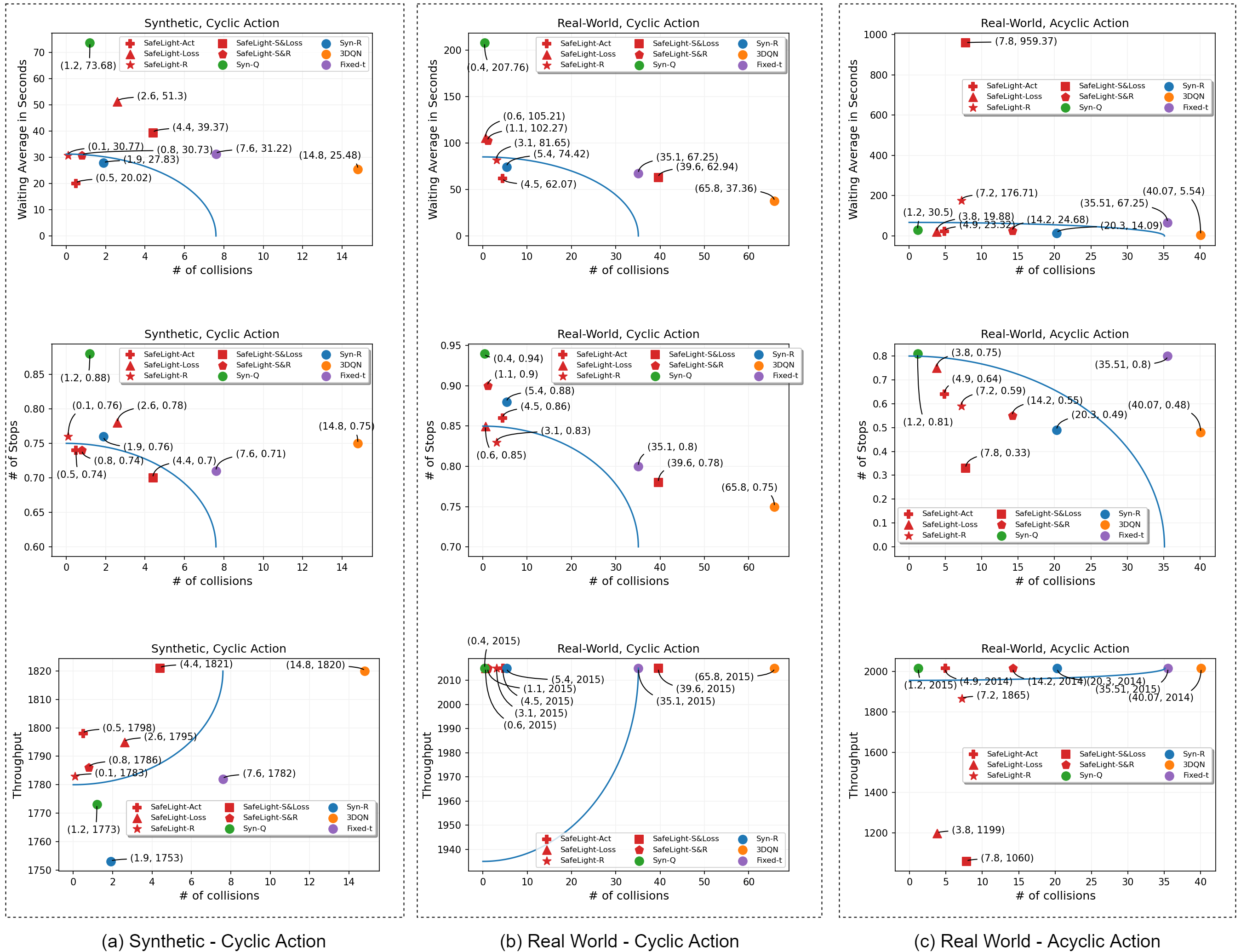}
  \caption{Performance Evaluation w.r.t Traffic Mobility and Safety. The \Safelight methods are marked in red and other baseline methods are marked in other colors as circles. The blue line in each figure encloses the best performance area. At least 3 \Safelight methods are within the best performance area in all 3 settings.}
  \label{fig:evaluation_performancce}
\end{figure*}

\subsubsection{Intersection Modeling}
We perform experiments on a synthetic intersection~\cite{Liang2019} and a real-world intersection~\cite{mei2022libsignal,ault2021reinforcement} with various settings. The synthetic intersection in Figure~\ref{fig:dataset}(a) has four approaches and twelve one-way vehicular movements. The real-world dataset is built based on the intersection at Cologne, Germany, as shown in Figure~\ref{fig:dataset}(b). Real traffic flow is adopted into the simulated environment. The intersection has 8 approaching lanes. In both intersections, the permitted left turn and U-turn are allowed. The setup of SUMO simulation environment assumes: (1) accurate vehicle detection, which is often available in practice via cameras or other sensors; (2) same deceleration rate, which can be modified based on needs and (3) permitted left turn in the timing plan.

\subsubsection{Collision Data} 
The collisions are generated by deliberately tuning up the probability that causes vehicles to ignore conflicting vehicles already in the intersection to mimic the aggressive driving behaviors, which are the root cause of any collisions~\cite{hamdar2008aggressiveness}. Using the generated collision data instead of the real-world collision data is more beneficial since (1) we can obtain a much larger collision dataset so that the sample is not biased; (2) unlike uncomfortable braking and predicted safety risk score, the number of collisions under this stress testing is the most straightforward indicator of safety hazards. 

\subsection{Comparison Methods}
We compare the performance of our approach with two RL methods that address safety under the same evaluation metrics to verify the effectiveness of our algorithm. Additionally, we compare against two baseline methods: the backbone RL model and the \formula controller. 
~\noindent\\$\bullet$
\formula~\cite{koonce2008traffic}: The lights are controlled solely on preset timings.
~\noindent\\$\bullet$
\singleobjective~\cite{Liang2019}: The backbone RL model minimizes delay without considering safety.
~\noindent\\$\bullet$
Synthetic Reward (\syntheticreward)~\cite{khamis2014adaptive}: This method uses a synthetic reward defined as $R(s,a) = w_1R_1(s,a) + w_2R_2(s,a)$. It includes two terms, each associated with an objective.
~\noindent\\$\bullet$
Synthetic Value (\syntheticvalue)~\cite{Gong2020}: The linearly weighted sum of Q-values for the two objectives is used to obtain a synthetic Q function.

\begin{figure*}[!t]
  \centering
  \includegraphics[width=.99\textwidth]{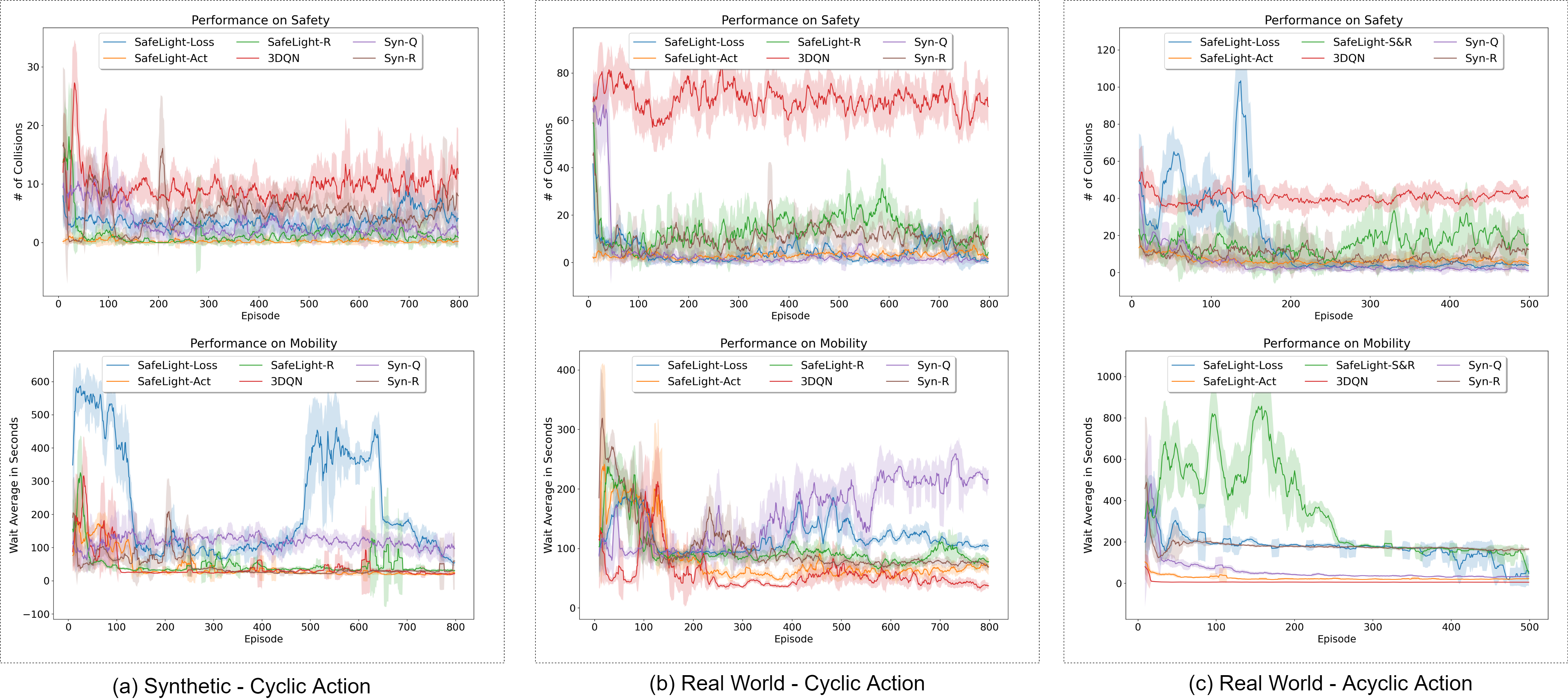}
  \caption{The training processes of \Safelight and other baseline methods under 3 different environment settings. The best performance considering both safety and mobility in 3 settings are \SafelightR, \SafelightAct, and \SafelightAct, respectively. The fastest to reach optimum in 3 settings are \SafelightR, \SafelightLoss, and \SafelightAct, respectively.}
  \label{fig:training_performancce}
\end{figure*}

\subsection{Experiment Result}
Here, we present the experimental results on two benchmark datasets (synthetic and real-world), two RL designs (cyclic and acyclic action spaces), and two backbone RL models (3DQN and PPO) to verify the effectiveness and generalizability of our proposed \Safelight. Both evaluation and training results are shown.

\subsubsection{Evaluation Metric}
We evaluate the performance on both mobility and safety and answer the above research questions. The average waiting time, throughput, and the number of stop vehicles are chosen as the performance measure for mobility. The number of collisions that happened within one simulation time frame is selected as the safety measure.

\subsubsection{Results of Evaluation Performance (\textbf{RQ1})}
Figure \ref{fig:evaluation_performancce} summarizes the key performance under three different settings (synthetic dataset and cyclic action space, real-world dataset and cyclic action space, real-world dataset and acyclic action space). We have the following observations:

\noindent$\bullet$ Though maintaining relatively better mobility, \singleobjective has the highest rate of collisions, even worse than the \formula control method in all settings, which verifies our viewpoint that \emph{existing RL-based work imposes safety issues and thus needs safety design}.
\syntheticvalue is better at reducing collisions but increases the average waiting time compared with \singleobjective.

\noindent$\bullet$ All the \Safelight methods consistently outperform \singleobjective in reducing collisions, while showing different performances in reducing the waiting time. \SafelightAct achieves 93.94\% and 86\% fewer collisions on synthetic and real-world datasets, respectively.
\SafelightR outperforms others on the synthetic setting in terms of the collision rates, with a 99.32\% improvement in safety. 

\noindent $\bullet$ We also noticed that embedding the safety model into the state does not seem to bring a boost in performance. This is similar to supervised learning, in which adding an additional feature without adding any labeling information will not guide the model learning.

\noindent $\bullet$ We especially paid attention to whether a safe action now may lead to future unsafe action in \SafelightAct, since intervention is made when the action of RL agent is determined as unsafe. Through prudent analysis, we found there are two cases in the consecutive states after the intervention: no opposing through movement vehicles or endless ones. In the former case, left-turn drivers will grant the right of the way. In the latter case, left turns are always prohibited, and left-turn drivers will have the green in the subsequent protected left-turn phase. Neither of them will result in an unsafe state.

\noindent $\bullet$ We observed  the cases where \SafelightR avoided collisions while \SafelightAct caused collisions. Although by construction, \SafelightAct should be able to prevent all the collisions, through collision visualization, we found collisions occurred in one case where the left-turn vehicle was entering the intersection during the green light and the opposing through vehicle was also driving in during the yellow light. In another case, a vehicle entered the intersection during the yellow light and was still in it when the yellow was switched to red; thus, the opposing vehicle conflicted with that vehicle. The \SafelightAct cannot eliminate the above two cases (i.e., a collision involving a yellow phase).

\subsubsection{Convergence during Training}
The convergence curves of RL methods are shown in Figure~\ref{fig:training_performancce}. 
Compared with \singleobjective, \SafelightAct can quickly converge to lower average waiting times and maintain strong safety performance throughout the training. This is reasonable since this method directly corrects the unsafe learned action. Although the reward is delayed under \SafelightAct, from the results we can still observe that occasional intervention does not hurt the learning much.
\begin{table}[h]
\centering
\begin{tabular}{c|c|p{4em}|p{3em}}
\hline
\multirow{2}{*}{} & & Mobility & Safety\\
 \cline{3-4}
Backbone RL & Methods & Average Delay & collisions \\
\hline
\multirow{7}{*}{}
         &\SafelightAct& \hfil16.57 & \hfil0.2\\
         &\SafelightLoss& \hfil13.07 & \hfil3.8\\
          &\SafelightR& \hfil16.60 & \hfil\textbf{0.0}\\
         \cline{2-4}
        \textit{IPPO} &\syntheticreward& \hfil20.02&\hfil0.8\\
         &\syntheticvalue& \hfil25.47 & \hfil7.3\\
         \cline{2-4}
         &\emph{3DQN}& \hfil\textbf{10.66} & \hfil24.5\\
         \cline{2-4}
          &\formula& \hfil27.38&\hfil7.0\\

\hline
\end{tabular}
\caption{Evaluation performance on \emph{IPPO}. Evaluation result is based on average value of 10 runs.}
\label{tab:ippo_evaluation}
\end{table}

\subsubsection{Study of Action Generalizability (RQ2)}
As discussed in earlier sections, there are two types of action spaces in TSC, i.e., cyclic and acyclic actions. Here, we investigate whether our design could generalize to different action settings.

In general, the acyclic action space increases collisions, which is one of the reasons why most existing real-world traffic signals operate in a cyclic manner. Among all the methods, \SafelightLoss and \SafelightAct perform relatively well under both the cyclic and acyclic action settings. In comparison, the \emph{Syn-R} can perform well in the cyclic setting but fail to minimize collisions in the acyclic setting. The performance gap in different settings attributes to the ad-hoc tuning of two reward factors, which verifies that the performance is sensitive to individual settings.

\subsubsection{Study of Model Generalizability {(RQ3)}} 
To verify that the proposed \Safelight can maintain superior performance under different backbone RL models. We adopt another RL model, \emph{IPPO}, from the benchmark~\cite{ault2021reinforcement} to evaluate the performance in the same synthetic environment as the previous experiment. The state space is defined as the queue length plus the max waiting time per lane. The action is defined as acyclic. The reward function is defined as the same as the state space. Table~\ref{tab:ippo_evaluation} shows the evaluation results. We can see that \Safelight can still outperform other baseline methods in terms of minimizing collisions.

\section{Conclusion}
In this work, we address the safety concern of current RL methods. By exploiting domain knowledge on safety, we propose a safety-enhanced reinforcement learning method that integrates a safety module into the underlying RL model. Our proposed method is compared with other baseline models under various benchmark datasets to show its effectiveness. We demonstrate that our model also has good generalizability under different settings, which is critical for real-world deployment with constantly changing traffic demands. We share our pathway for finding a way to enforce safety which can be a guideline for future study.

Our proposed model could be utilized for more specific domain safety rules, including yellow and red timing. In the future, we will extend our method to more transportation rules toward a safer RL-based TSC method.

\section*{Acknowledgments}
This research was supported, in part, by Federal Highway Administration's Exploratory Advanced Research (EAR) Program under Grant Number 693JJ320C000021, and by National Science Foundation (USA) under Grant Numbers CNS--1948457 and IIS-2153311. The views and conclusions contained in this paper are those of the authors and should not be interpreted as representing any funding agencies.

\bibliography{aaai23}
\end{document}